\documentclass[journal]{IEEEtran}
\usepackage{amsmath,amssymb,bm}
\usepackage{booktabs}
\usepackage{graphicx}
\usepackage{array}
\usepackage{multirow}
\usepackage{textcomp}
\usepackage{threeparttable}
\usepackage[hidelinks]{hyperref}
\usepackage{microtype}
\usepackage{orcidlink}

\graphicspath{{figs/}}

\newcommand{\A}{\mathcal{A}}
\newcommand{\D}{\mathcal{D}}
\newcommand{\Hh}{\mathcal{H}}

\begin{document}

\title{HB-PVI: A Hierarchical Bayesian Personalization and Value-of-Information Framework for Complex Activity Recognition}

\author{Hammed A. Olayinka~\orcidlink{0000-0002-9796-5276}
\thanks{H. A. Olayinka is with the Department of Mathematical Sciences, Worcester Polytechnic Institute, Worcester, MA 01609 USA (e-mail: haolayinka@wpi.edu).}}

%\markboth{IEEE JOURNAL OF BIOMEDICAL AND HEALTH INFORMATICS}{Olayinka: HB-PVI}

\maketitle

\begin{abstract}
Personalization can improve activity-recognition performance, but participant-specific gains are heterogeneous and every additional calibration label has an acquisition cost. This study presents HB-PVI, a hierarchical Bayesian personalization and value-of-information framework that jointly models participant heterogeneity, the benefit and harm of four personalization mechanisms, and the economic value of an additional label, for the MUSIC-CAR complex-activity cohort of 47 participants. A leakage-safe, leave-one-participant-out evaluation combines a sequential-Monte-Carlo participant-effect updater with a Student-$t$ hierarchical gain model and a one-step expected-value-of-sample-information (EVSI) stopping rule. Adapter personalization produced small positive mean F1 gains that grew from 0.00099 at one label to 0.00198 at ten labels, whereas joint adapter-plus-head and prototype-residual personalization were negative on average. Under the primary practical-benefit threshold ($\Delta_{\min}=0.01$) and cost setting, one-step EVSI was exactly zero at every decision state, so the policy purchased no labels and retained population inference for all 47 participants, matching always-stop exactly (region-of-practical-equivalence probability $=1$). Relative to fixed ten-shot adapter personalization, this reduced labeling by 100\% while keeping the posterior mean F1 loss at 0.00217 (95\% credible interval, 0.00048 to 0.00389), with posterior probability 0.9992 of remaining below the 0.005 tolerance. HB-PVI was utility-optimal in 199 of 216 cost-threshold settings and in every setting at or above the primary label cost. These results argue for a population-first deployment policy whenever personalization gains are small relative to labeling, computation, and harm costs, and they show why value-of-information reasoning, not raw predictive accuracy, should drive personalization decisions in health-sensing applications.
\end{abstract}


\section{Introduction}

Passive monitoring of complex, health-indicative activities of daily living (ADLs) from smartphone and wearable sensors offers a scalable alternative to infrequent, rater-dependent clinical assessment \cite{phart}. Wearable-sensor human activity recognition (HAR) has matured substantially over the last decade \cite{bulling2014}, yet unlike simple ambulatory activities, complex activities (CAs) are composed of simple activities performed sequentially, concurrently, or interleaved \cite{ranasinghe2016}, and the same CA can be executed in markedly different ways by different people \cite{musiccar}. Population models trained across many participants therefore often generalize poorly to a specific new user \cite{phart}, motivating \emph{personalization}: adapting a population model with a small number of participant-specific labels. Personalization is not free, however. Adapting on unrepresentative or noisy calibration windows can degrade performance for participants whom the population model already served well \cite{phart}, and every calibration label carries an acquisition cost in patient or clinician time. A deployable personalization policy must therefore decide, for each participant and at each point in a calibration sequence, \emph{which} personalization mechanism (if any) to apply and \emph{whether} to purchase another label at all.

Prior work on this cohort established that a personalized transformer (P-HART) with fixed 6-shot user adapters raised mean F1 from 88.0\% to 92.6\% on average, though personalization improved performance for 41 of 47 participants and slightly reduced it for the remaining six, who already had high non-personalized performance \cite{phart}, and that complex-activity recognition itself is substantially harder than simple-activity recognition, with baseline machine-learning F1 of only 58.8\% on the underlying MUSIC-CAR dataset \cite{musiccar}. Neither study modeled participant-level uncertainty in the personalization decision itself or priced the value of an additional calibration label against its cost. This paper addresses this gap with HB-PVI (Hierarchical Bayesian Personalization with Value of Information): a fully Bayesian inferential and decision layer, conditioned on deterministic cross-fitted sensor representations, that (1) propagates population and participant uncertainty into calibrated activity predictions, (2) models the participant-specific benefit and harm distribution of four personalization mechanisms, (3) selects the mechanism maximizing posterior expected utility net of labeling, computation, and harm costs, and (4) uses a one-step expected value of sample information (EVSI) rule \cite{howard1966,raiffa1961,chaloner1995} to decide whether a further label is worth acquiring. Framing personalization as a sequential label-purchase decision, rather than a one-shot model-selection problem, distinguishes HB-PVI from few-shot and meta-learning approaches to personalization \cite{finn2017} that optimize predictive accuracy without pricing the cost of the adaptation data itself.

HB-PVI builds on a substantial Bayesian and decision-theoretic literature. Hierarchical partial pooling and regularized shrinkage priors \cite{gelman2013,piironen2017} let participant-level effects borrow strength across a modest 47-participant cohort; Hamiltonian Monte Carlo with the No-U-Turn Sampler \cite{hoffman2014,betancourt2017,carpenter2017} provides asymptotically exact posterior inference rather than a variational or Laplace approximation \cite{lakshminarayanan2017,daxberger2021}; and value-of-information and Bayesian experimental-design theory \cite{howard1966,chaloner1995,chick2017,settles2009} formalizes when to stop collecting labels, echoing expected value of perfect/sample information (EVPI/EVSI) methods long used in medical decision-making and health-technology assessment \cite{felli1998,ades2004}, applied here at the level of an individual participant's calibration sequence rather than a population-level trial design. This study uses the region-of-practical-equivalence (ROPE) framework \cite{kruschke2018} to test practical equivalence to simpler comparators, and adopts convergence and calibration diagnostics standard in applied Bayesian workflows \cite{vehtari2021,talts2018}. The personalization mechanisms (lightweight adapters, classification-head fine-tuning, and their combination) parallel parameter-efficient transfer learning \cite{houlsby2019,pan2010} applied to the attention-based sequence backbone used for complex-activity recognition \cite{vaswani2017,cho2014,phart}.

HB-PVI also sits alongside just-in-time adaptive interventions and contextual-bandit algorithms in mobile health, which decide in real time which intervention to deliver given a person's current context \cite{nahumshani2018,tewari2017}; that literature typically treats context as free, whereas HB-PVI addresses the upstream question of whether the context itself is worth purchasing.

This study's contributions are: (i) a fully Bayesian personalization-benefit model that reports calibrated, participant-specific probabilities of meaningful benefit and harm for four personalization mechanisms; (ii) a validated sequential-Monte-Carlo participant-effect updater that maintains at least 400 distinct population-posterior ancestors while incorporating adaptation labels one at a time; (iii) a one-step EVSI stopping rule, evaluated over a sensitivity grid of 216 practical-benefit-threshold and cost combinations, that determines whether personalization is worth its price; and (iv) a complete leakage-safe evaluation on 47 participants showing that, under primary costs, no further labels are justified and a population-first policy preserves F1 within a primary tolerance while eliminating calibration labeling entirely.

\section{Methods}

\subsection{Cohort, design, and leakage control}
This study used the MUSIC-CAR cohort of 47 participants performing sequential, concurrent, and interleaved complex ADLs (nine activity labels, $k=1,\dots,9$) from wrist- and pocket-mounted smartphone sensors \cite{musiccar}. The outer evaluation followed leave-one-participant-out logic: for held-out participant $p^\star$, all representation training, normalization, and hyperparameter selection excluded that participant. Windows were ordered prospectively into a 10-window adaptation block, two 12-window guard gaps, a 128-window calibration block, and a prospective test block, so a decision at shot $n$ ($n=0,\dots,10$) used only the first $n$ purchased labels. Because adjacent 60-second, 5-second-stride windows overlap by approximately 92\%, the primary likelihood used a deterministic, label-blind non-overlapping subsample to limit pseudo-replication.

\subsection{Deterministic representation and cross-fitting}
A nine-member deep ensemble of P-HART-style backbones \cite{phart} (an attention encoder \cite{vaswani2017} with a bidirectional GRU \cite{cho2014} temporal head) was trained on nested-training participants only, with fold-specific normalization and training-only principal-components analysis (PCA). One deterministic member supplied the $q$-dimensional pooled representation $\mathbf z_{pi}\in\mathbb R^q$ (participant $p$, window $i$) used to condition every downstream Bayesian model, so a single, fixed latent coordinate system was shared across all analyses; the remaining members were used only to compute the predictive-uncertainty summaries (mutual information, entropy, predictive variance) entering the Model II sequential state (Section~\ref{sec:modelii}), not as an approximate posterior. Every unknown quantity downstream of $\mathbf z_{pi}$ was assigned a prior and sampled with Hamiltonian Monte Carlo / NUTS \cite{hoffman2014,carpenter2017,betancourt2017}, so inference is fully Bayesian conditional on leakage-safe representations, without variational or Laplace approximation \cite{lakshminarayanan2017,daxberger2021}.

\subsection{Model I: hierarchical multi-label prediction and sequential updating}
For participant $p$, window $i$, and activity $k$, let $y_{pik}\in\{0,1\}$ denote the observed label. Conditional on the representation $\mathbf z_{pi}$,
\begin{equation}
y_{pik}\sim\mathrm{Bernoulli}(\sigma(\eta_{pik})),\quad \eta_{pik}=\alpha_k+\mathbf z_{pi}^{\mathsf T}\bm\beta_k+b_{pk},
\end{equation}
where $\sigma(\cdot)$ is the logistic function; $\alpha_k$ is an activity-specific intercept; $\bm\beta_k$ is an activity-specific representation-coefficient vector; and $b_{pk}$ is a participant-specific intercept for activity $k$, collected across activities as $\mathbf b_p=(b_{p1},\dots,b_{p9})^{\mathsf T}$. Because the $q$-dimensional representation is high-dimensional relative to the per-fold training data, each entry of $\bm\beta_k$ follows a regularized horseshoe prior \cite{piironen2017} with activity-specific local and global shrinkage scales and a finite slab scale of 0.5 (full specification in Supplementary Section~S1); the fitted primary specification used this horseshoe on $\bm\beta_k$ alone, with no additional residual cross-activity factor term. Participant intercepts follow $\mathbf b_p\sim\mathcal N_9(\mathbf 0,\operatorname{diag}(\bm\tau_b)\,\bm\Omega_b\,\operatorname{diag}(\bm\tau_b))$, where $\bm\tau_b$ are activity-specific participant-effect scales and $\bm\Omega_b$ is a correlation matrix with an LKJ prior \cite{lkj2009}. Let $\bm\Theta$ collect all population-level parameters ($\alpha_k$, $\bm\beta_k$, $\bm\tau_b$, $\bm\Omega_b$, and the analogous Model II parameters introduced below). For an unseen participant $p^\star$, after $n$ purchased adaptation windows $\A_{p^\star,n}$ (the labels from the first $n$ chronological adaptation windows), and writing $\D_{-p^\star}$ for all training-side data excluding participant $p^\star$, the participant-effect posterior is
\begin{align}
&p(\mathbf b_{p^\star},\bm\Theta\mid \D_{-p^\star},\A_{p^\star,n})\notag\\
&\quad\propto p(\A_{p^\star,n}\mid\mathbf b_{p^\star},\bm\Theta)\,p(\mathbf b_{p^\star}\mid\bm\Theta)\,p(\bm\Theta\mid\D_{-p^\star}).
\end{align}
This posterior is maintained sequentially, one purchased label at a time, by an adaptive tempered resample-move sequential Monte Carlo algorithm: each of 16,000 retained population-posterior draws seeds four conditionally independent participant-effect particles (64,000 joint particles); systematic resampling triggers only when effective sample size (ESS) falls below 25\% of 64,000; random-walk Metropolis-Hastings rejuvenates the nine-dimensional participant effect after every resampling step, with incremental likelihoods recomputed post-rejuvenation. Because population-level parameters are never rejuvenated, every retained state was required to preserve at least 400 distinct population-posterior ancestors in addition to a final particle ESS of at least 400; all 47 folds passed both gates (full algorithm and validation-gate detail in Supplementary Section~S1).

\subsection{Personalization actions, gain, and Model II}
\label{sec:modelii}
Four personalization mechanisms $a\in\A$ =\{\text{adapter}, \text{head}, \text{adapter+head}, \text{prototype-residual}\} were compared against a zero-gain population baseline \cite{houlsby2019}: the adapter mechanism fine-tunes a small low-rank bottleneck inserted into the backbone; the head mechanism fine-tunes only the final classification layer; adapter-plus-head fine-tunes both jointly (roughly doubling the number of participant-adapted parameters relative to either mechanism alone); and prototype-residual personalization adjusts predictions using a participant-specific residual from a nearest-prototype reference, without gradient fine-tuning. For action $a$, shot $n$, and participant $p$, gain is $g_{pan}=F1_{pan}^{a}-F1_{p}^{\mathrm{population}}$ (sample-level F1 for action $a$ minus the population-only baseline for the same participant), evaluated on a held-out set of $47\times4\times10=1{,}880$ fold-action-shot observations. A robust hierarchical benefit model was fit as
\begin{align}
100\,g_{i} &\sim t_{\nu}(\alpha_{a_i}+c_{p_ia_i}+\mathbf x_i^{\mathsf T}\bm\gamma_{a_i},\ \sigma_{a_i}),\\
\mathbf c_p &\sim \mathcal N(\mathbf 0,\operatorname{diag}(\bm\tau)\,\bm\Omega\,\operatorname{diag}(\bm\tau)),
\end{align}
where $i$ indexes a fold-action-shot observation with action $a_i$ and participant $p_i$; $\alpha_{a_i}$ is an action-specific intercept; $\mathbf x_i$ is the corresponding sequential-state predictor vector (defined below); $\bm\gamma_{a_i}$ is an action-specific coefficient vector (distinct from Model I's $\bm\beta_k$), assigned a fixed-scale, weakly informative Gaussian prior rather than a sparsity-inducing one, since with only 19 predictors this was judged sufficient (unlike Model I's higher-dimensional $\bm\beta_k$; full priors for both models in Supplementary Section~S1); $c_{pa}$ is a participant-by-action random intercept, collected as $\mathbf c_p=(c_{p,\text{adapter}},\dots)^{\mathsf T}$ with scales $\bm\tau$ and correlation matrix $\bm\Omega$; $\sigma_{a_i}$ is an action-specific residual scale; and $\nu$ is the Student-$t$ degrees of freedom (bounded below at 2 for finite variance), which downweights outlying participant-fold gains relative to a Gaussian likelihood. The predictor vector $\mathbf x_i$ contains 19 sequential-state predictors spanning purchased-label prevalence and diversity, ensemble uncertainty, posterior participant-effect summaries, calibration probability, predictive entropy, shot-to-shot change, and representation shift (Supplementary Table~S3). Action-specific coefficient vectors allow each state predictor to have a different association with gain for each personalization mechanism; Model II did not use a horseshoe prior or an additional interaction layer beyond these action-specific coefficients. Four chains were fit per held-out fold with convergence gates (rank-normalized $\widehat R<1.01$ \cite{vehtari2021}, zero divergences, minimum bulk/tail ESS $>400$); all 47 folds passed with larger ESS $(>1000)$ and smaller $\widehat R<1.01$.

\subsection{Utility, Bayes action, and EVSI}
Let $\Hh_n$ denote the sequential state available after $n$ purchased labels (the posterior summaries and purchased-label history described above). Cumulative realized utility for action $a$ after $n$ labels with gain $g$ is
\begin{equation}
U(a,n,g)=g-nc_L-c_C(a)-c_R\,\mathbb I(g<0),
\end{equation}
where $c_L$ is the per-label acquisition cost, $c_C(a)$ is the action-specific computation cost, $c_R$ is a harm penalty applied when realized gain is negative, and $\mathbb I(\cdot)$ is the indicator function. The Bayes action maximizes posterior expected utility, $a_n^\star=\arg\max_a\mathbb E[U(a,n,g)\mid\Hh_n]$, subject to the conservative eligibility rule $P(g>\Delta_{\min}\mid \Hh_n)\geq q$ and $P(g<0\mid\Hh_n)\leq r$, where $\Delta_{\min}$ is the minimum practically meaningful gain, and $q=0.90$, $r=0.10$ are the primary minimum-benefit and maximum-harm probabilities. To avoid charging label cost twice when comparing the value of one more label, EVSI uses the cost-free utility $U'(a,g)=g-c_C(a)-c_R\,\mathbb I(g<0)$; writing $Y_{n+1}$ for the hypothetical outcome of the next purchased label,
\begin{align}
\mathrm{EVSI}_{n+1}=&\ \mathbb E_{Y_{n+1}\mid \Hh_n}\Big[\max_a\mathbb E\{U'(a,g)\mid \Hh_n,Y_{n+1}\}\Big]\notag\\
&-\max_a\mathbb E\{U'(a,g)\mid \Hh_n\}.
\end{align}
Acquisition continues only while $\mathrm{EVSI}_{n+1}>c_L$, evaluated with 256 hypothetical next-label draws per state along the fixed chronological window order; active (non-chronological) window selection was deferred to future work. The sensitivity analysis combined $\Delta_{\min}\in\{0,0.005,0.01\}$, $c_L\in\{0,0.0005,0.001,0.0025,0.005,0.01\}$, $c_C\in\{0,0.0005,0.001\}$, and $c_R\in\{0,0.005,0.01,0.02\}$ (72 cost combinations $\times$ 3 thresholds $=216$ settings), with primary values $c_L=0.001$, $c_C=0.0005$, $c_R=0.01$.

\subsection{Statistical analysis and study objectives}
Participant-level paired utility differences between HB-PVI and each comparator were modeled with Student-$t$ posteriors, reporting the posterior mean paired difference $\mu_d$, 95\% credible intervals, $P(\mu_d>0)$, and region-of-practical-equivalence probabilities $P(|\mu_d|<0.01)$ \cite{kruschke2018}. Held-out prediction was summarized by mean absolute error (MAE), root-mean-squared error (RMSE), proper Brier scores \cite{gneiting2007}, and 95\% predictive-interval coverage using five-bin reliability curves (bins with $<10$ observations omitted). The label-efficiency analysis evaluated whether HB-PVI reduced label use while maintaining a posterior probability greater than 0.90 that mean F1 loss was below 0.005. This analysis used both a participant-level Bayesian Student-$t$ model and an independent, non-parametric 200{,}000-resample participant bootstrap \cite{efron1993} to assess sensitivity to the likelihood. Detailed analytical objectives are reported in Supplementary Table~S12.

\section{Results}

\subsection{Action-specific personalization outcomes}
\begin{figure*}[!t]
\centering
\includegraphics[width=0.88\textwidth]{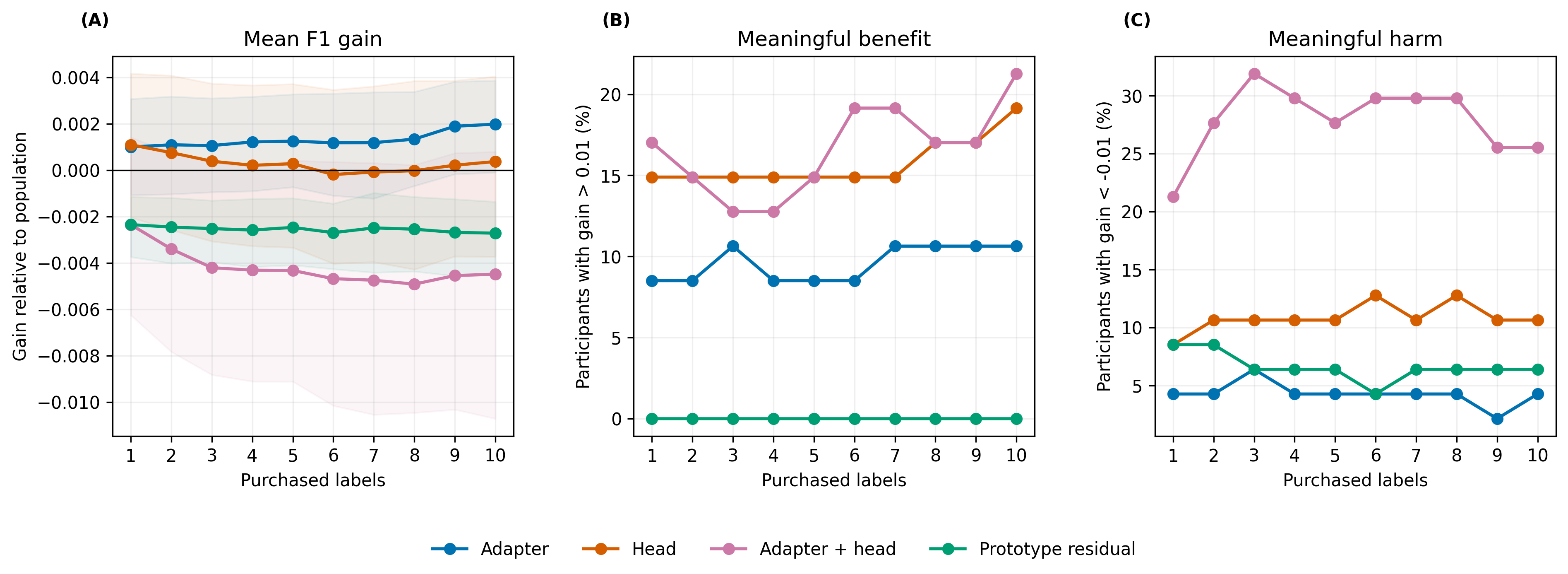}
\caption{Action-specific personalization outcomes across purchased labels. (A) Mean sample-level F1 gain relative to population inference, with 95\% intervals. (B) Percentage of participants with meaningful benefit ($g>0.01$). (C) Percentage with meaningful harm ($g<-0.01$).}
\label{fig:actions}
\end{figure*}

Table~\ref{tab:actions} summarizes gain at the first and tenth purchased label for each mechanism (full ten-shot trajectories in Fig.~\ref{fig:actions}). Adapter personalization was the only mechanism with a positive mean gain at every shot, rising from $+0.00099$ to $+0.00198$; 72.3\% of participants had positive adapter gain at 10 labels, with meaningful benefit ($g>0.01$) in 10.6\% and meaningful harm ($g<-0.01$) in 4.3\%. Head personalization was near zero on average and became more heterogeneous with more labels (meaningful benefit rising from 14.9\% to 19.1\%, meaningful harm from 8.5\% to 10.6\%). Adapter-plus-head and prototype-residual personalization were negative on average at every shot, with adapter-plus-head showing both the largest benefit subgroup (up to 21.3\%) and the largest harm subgroup (up to 31.9\%) of any mechanism.

\begin{table}[!t]
\caption{Action-specific personalization gain at 1 and 10 purchased labels (percentages of 47 participants)}
\label{tab:actions}
\centering
\scriptsize
\begin{threeparttable}
\begin{tabular}{lccccc}
\toprule
Action & Shot & Mean gain & Pos.\ (\%) & Ben.\ (\%) & Harm (\%)\\
\midrule
\multirow{2}{*}{Adapter} & 1 & $+0.00099$ & 61.7 & 8.5 & 4.3\\
 & 10 & $+0.00198$ & 72.3 & 10.6 & 4.3\\
\multirow{2}{*}{Head} & 1 & $+0.00109$ & 51.1 & 14.9 & 8.5\\
 & 10 & $+0.00037$ & 55.3 & 19.1 & 10.6\\
\multirow{2}{*}{Adapter+head} & 1 & $-0.00236$ & 44.7 & 17.0 & 21.3\\
 & 10 & $-0.00449$ & 48.9 & 21.3 & 25.5\\
\multirow{2}{*}{Prototype-resid.} & 1 & $-0.00235$ & 17.0 & 0.0 & 8.5\\
 & 10 & $-0.00272$ & 19.1 & 0.0 & 6.4\\
\bottomrule
\end{tabular}
\begin{tablenotes}[flushleft]
\footnotesize
\item[] Gain, sample-level F1 for the personalization action minus population-only F1; Pos., percentage of participants with gain $>0$; Ben., percentage with gain exceeding the meaningful-benefit threshold ($>0.01$); Harm, percentage below the meaningful-harm threshold ($<-0.01$).
\end{tablenotes}
\end{threeparttable}
\end{table}

\subsection{Population baseline strength and the ceiling on personalization value}
\label{sec:baseline}
\begin{figure*}[!t]
\centering
\includegraphics[width=0.88\textwidth]{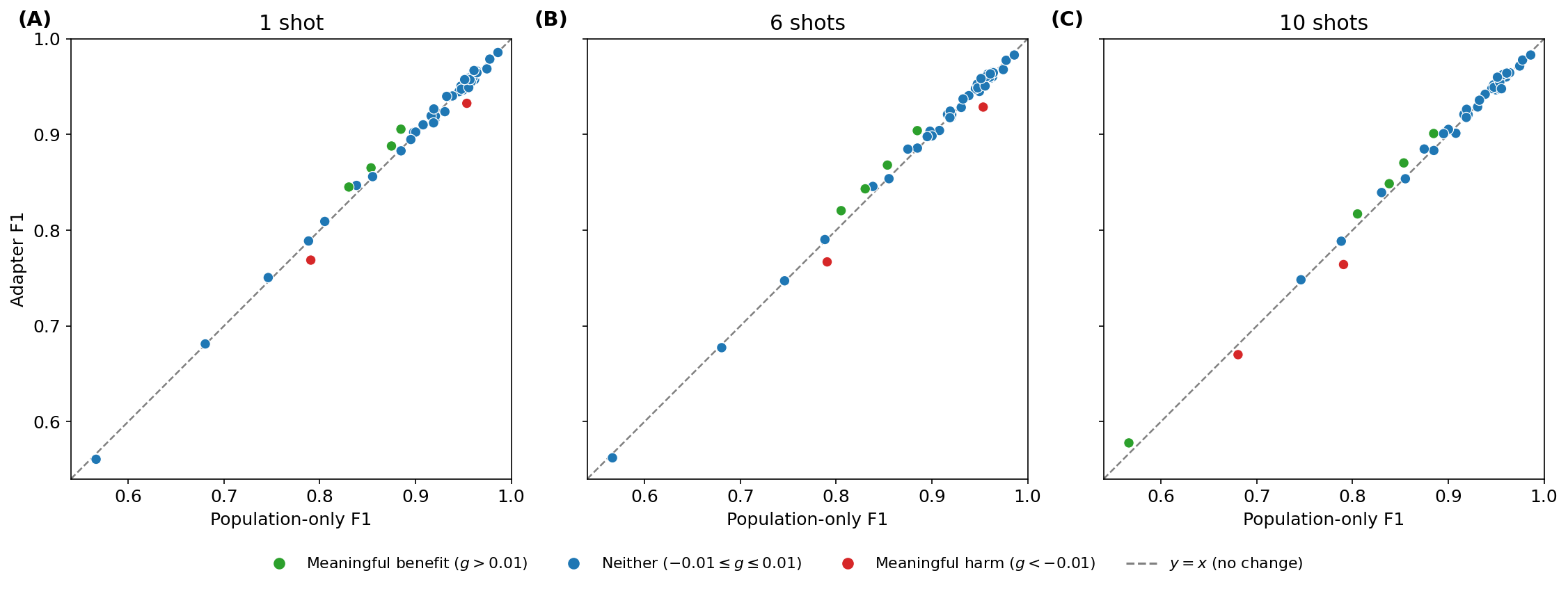}
\caption{Population-only F1 versus adapter F1 at (A) 1, (B) 6, and (C) 10 purchased labels, all 47 participants (Table~\ref{tab:perparticipant}). Points scatter closely around the $y=x$ line at every shot count, and meaningful-benefit (green) and meaningful-harm (red) cases both occur at low and high baseline F1 alike, illustrating that realized gain is not predictable from population-only performance ($r=-0.06$ at 10 shots.}
\label{fig:calibration}
\end{figure*}

The population-only ensemble member (no participant-specific label) achieved a mean held-out sample-level F1 of 0.9042 across the 47 participants -- the reference against which every gain in Table~\ref{tab:actions} and Fig.~\ref{fig:actions} is measured. This figure is informative when read against two related studies that share this cohort or sensor modality (Table~\ref{tab:crossstudy}). On the same MUSIC-CAR data, five traditional feature-based machine-learning baselines reached a best mean F1 of only 58.8\% \cite{musiccar}, underscoring that complex-activity recognition from raw statistical features is substantially harder than the simple-activity recognition problem those baselines were designed for. On the same 47-participant cohort, a non-personalized P-HART transformer reached F1 of 88.0\%, rising to 92.6\% after few-shot user-adapter personalization \cite{phart}. The population-only backbone used by HB-PVI exceeded the previously reported non-personalized P-HART result by approximately 0.024 F1, narrowing, but not eliminating, the 0.046 F1 gap between the previously reported non-personalized and personalized P-HART results. Because the studies differed in training, ensemble configuration, windowing, and evaluation protocol, this comparison is contextual rather than controlled. Fig.~\ref{fig:calibration} plots each participant's population-only F1 against adapter F1 at 1, 6, and 10 labels; most points sit near $y=x$, and a few participants account for most of the departure. The lowest-baseline participant (F1$=0.566$) shows neither benefit nor harm at 1 or 6 shots but crosses into meaningful benefit by 10, showing classification can still change late in the sequence. The largest-harm participant (F1$\approx0.79$) remains meaningful harm at all three shots, so not every unfavorable case is transient.

\begin{table}[!t]
\caption{Population and personalized mean F1 across related studies on the same cohort or sensor modality}
\label{tab:crossstudy}
\centering
\scriptsize
\begin{threeparttable}
\begin{tabular}{lccc}
\toprule
Study & Backbone & Personalization & Mean F1\\
\midrule
MUSIC-CAR baseline \cite{musiccar} & Feature+XGBoost & None & 0.588\\
P-HART \cite{phart} & P-HART transf. & None & 0.880\\
P-HART \cite{phart} & P-HART transf. & 6-shot adapter & 0.926\\
HB-PVI (this study) & P-HART-style\tnote{a} & None (population) & 0.904\\
HB-PVI (this study) & P-HART-style\tnote{a} & Adapter, 6 shots & 0.905\\
HB-PVI (this study) & P-HART-style\tnote{a} & Adapter, 10 shots & 0.906\\
\bottomrule
\end{tabular}
\begin{tablenotes}[flushleft]
\footnotesize
\item[a] HB-PVI used a P-HART-style backbone within a nine-model deep ensemble; one deterministic member supplied the representations used in the downstream Bayesian analyses, while the remaining members contributed predictive-uncertainty summaries. The MUSIC-CAR and P-HART F1 values were taken from the cited publications and were not recomputed here; they are provided for context only, not a controlled head-to-head comparison.
\end{tablenotes}
\end{threeparttable}
\end{table}

Table~\ref{tab:crossstudy} should be read as context rather than a controlled comparison: the studies differ in ensemble configuration, contrastive-loss training, cross-validation protocol, and evaluation windowing. It nonetheless offers one candidate explanation for why the gains identified by the Model II gain model were both small in absolute F1 units and inconsistent in sign across mechanisms: when the population baseline already captures most of the recoverable signal, the marginal information content of a calibration label about \emph{which} action will help a specific participant is itself small -- precisely the condition under which one-step EVSI is expected to be low. This mean result concealed substantial participant-level heterogeneity: realized adapter gain ranged from $-0.0219$ to $+0.0208$ at 1 shot, $-0.0249$ to $+0.0193$ at 6 shots, and $-0.0265$ to $+0.0168$ at 10 shots (Supplementary Table~S6), even though the population-averaged gain never exceeded $+0.00198$. Exact participant-specific F1 scores are reported in Supplementary Table~S7.

\subsection{Held-out gain prediction and calibration}
\label{sec:calibration}
Table~\ref{tab:calibration} report held-out prediction accuracy. Prototype-residual gain was easiest to predict (MAE 0.00325, 95.3\% coverage) and adapter-plus-head hardest (MAE 0.01396, 88.9\% coverage); the two more aggressive mechanisms had both larger prediction error and below-nominal predictive coverage, supporting the conservative eligibility rule used for action selection. Harm-specific discrimination was weaker still: comparing each action's harm Brier score against a trivial baseline that always predicts the action's empirical harm rate, the fitted model was less accurate than this trivial baseline for every one of the four mechanisms (e.g., adapter: model 0.269 versus trivial 0.229; prototype-residual: model 0.234 versus trivial 0.165). Full reliability curves showing where each mechanism's calibration breaks down (rather than just the summary coverage and Brier statistics in Table~\ref{tab:calibration}) are provided in Supplementary Fig.~S1. The gain model therefore did not demonstrate prospective participant-level discrimination of personalization harm beyond the base rate (Section~\ref{sec:crit3}).

\begin{table}[!t]
\caption{Held-out gain-model prediction and calibration by action}
\label{tab:calibration}
\centering
\scriptsize
\begin{threeparttable}
\begin{tabular}{lccccc}
\toprule
Action & MAE & RMSE & Cov.\ (\%) & Brier$_{g>0}$ & Brier$_{g>.01}$\\
\midrule
Adapter & 0.00565 & 0.00824 & 91.9 & 0.2695 & 0.0915\\
Head & 0.00958 & 0.01321 & 91.1 & 0.2721 & 0.1257\\
Adapter+head & 0.01396 & 0.01857 & 88.9 & 0.2940 & 0.1318\\
Prototype-resid. & 0.00325 & 0.00546 & 95.3 & 0.1860 & 0.0002\\
\bottomrule
\end{tabular}
\begin{tablenotes}[flushleft]
\footnotesize
\item[] MAE, mean absolute error (F1 units); RMSE, root-mean-squared error (F1 units); Cov., empirical coverage of the nominal 95\% posterior predictive interval; Brier $g>0$ and $g>.01$, Brier scores for predicting positive gain and gain exceeding the meaningful-benefit threshold, respectively (lower is better; 0 is perfect).
\end{tablenotes}
\end{threeparttable}
\end{table}

\subsection{EVSI, stopping, and the primary deployment decision}
\begin{figure*}[!t]
\centering
\includegraphics[width=0.88\textwidth]{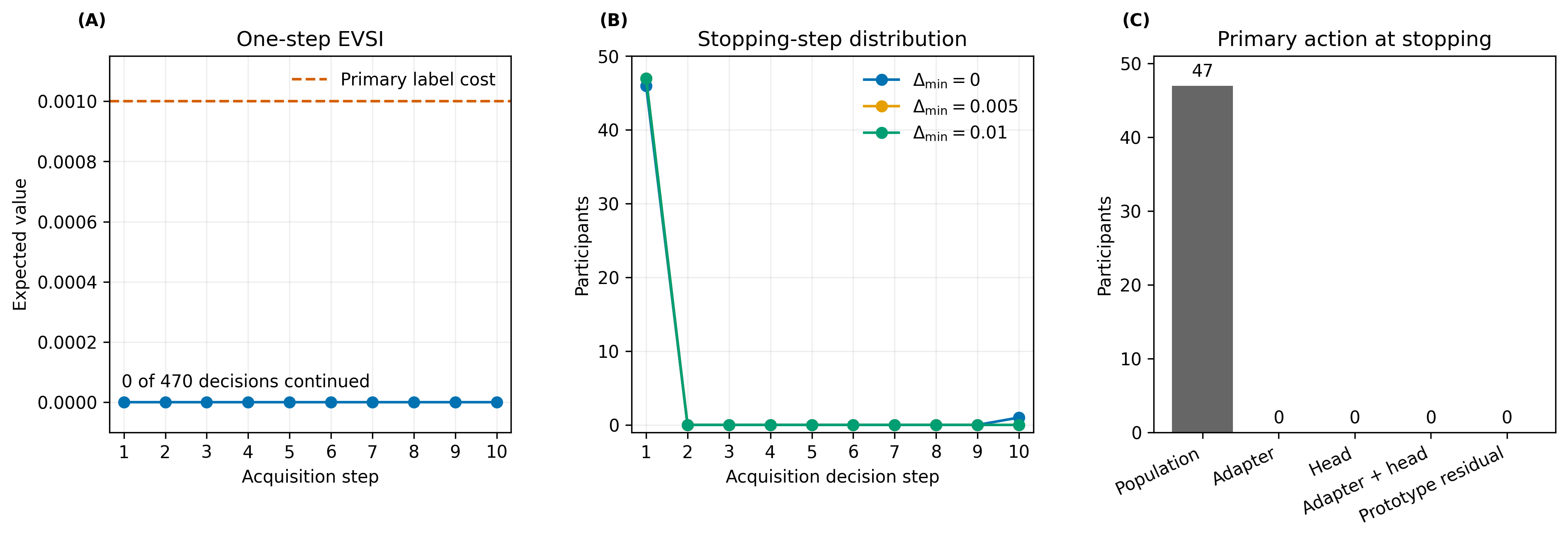}
\caption{Fixed-order one-step EVSI and stopping decisions. (A) Mean EVSI under the primary setting; no state-level decision continued. (B) Labels purchased across the three practical-gain thresholds. (C) Action selected at stopping under the primary $\Delta_{\min}=0.01$ setting.}
\label{fig:evsi}
\end{figure*}

Under the primary cost setting, one-step EVSI was exactly zero for every state transition from shot 1 through shot 10 (Fig.~\ref{fig:evsi}A), so none of the 470 participant-state decisions continued acquisition. At $\Delta_{\min}\in\{0.005,0.01\}$ the policy stopped immediately, purchased zero labels for all 47 participants (Fig.~\ref{fig:evsi}B), and retained population inference (Fig.~\ref{fig:evsi}C); realized mean gain and cumulative utility were both exactly zero and no participant was exposed to personalization harm. At $\Delta_{\min}=0$ the policy purchased a mean of 0.213 labels (95\% bootstrap interval 0.000--0.638), selecting head personalization for one of the 47 participants after 10 labels and population inference for the rest (mean gain $+0.000284$, mean utility $+0.000061$).

\subsection{One-step EVSI ceiling across the sensitivity grid}
\label{sec:evsiceiling}
Section III-C reports EVSI at the primary setting; here we characterize its largest observed value anywhere in the primary design. Across all 47 participants and 216 cost-threshold settings (10,152 policy decisions), a strictly positive one-step EVSI occurred in only 72 decisions (0.7\%), exclusively at $\Delta_{\min}=0$; at $\Delta_{\min}\in\{0.005,0.01\}$ no decision, at any label cost, had positive EVSI. All 72 positive-EVSI decisions involved the same single participant -- the one participant for whom head personalization was ultimately selected at $\Delta_{\min}=0$ (Section III-C). At label costs at or below the primary value ($c_L\leq0.001$), this participant's EVSI exceeded cost at every state, so the policy purchased labels through the full ten-shot horizon (mean EVSI per continued step, 0.0041). At higher label costs ($c_L\geq0.0025$), EVSI exceeded cost only immediately after the first label -- its peak value anywhere in the study, 0.01012, narrowly exceeding the largest primary label cost of 0.01 -- and then fell below cost, so the policy purchased exactly one label before stopping.

\subsection{Utility comparisons, paired differences, and cost sensitivity}
Table~\ref{tab:utility} combines realized comparator utility with the posterior paired-difference analysis. Fixed adapter personalization raised mean F1 but had negative cumulative utility once labeling, computation, and harm costs were charged (e.g., $-0.01129$ at 10 shots versus $+0.00198$ mean gain alone). Because HB-PVI selected population inference for every participant, it was exactly equivalent to always-stop (all 47 paired differences $=0$; ROPE probability $=1$). Relative to fixed adapter personalization, the posterior probability that HB-PVI's utility was strictly greater was $\geq0.989$ at every shot count tested and reached $1.0000$ at 6 and 10 shots.

\begin{table}[!t]
\caption{Realized comparator utility and posterior paired-utility difference (HB-PVI minus comparator)}
\label{tab:utility}
\centering
\scriptsize
\begin{threeparttable}
\setlength{\tabcolsep}{3.5pt}
\begin{tabular}{lccccc}
\toprule
Comparator & Util. & $\Delta$ mean & 95\% CrI & $P(\Delta\!>\!0)$\\
\midrule
Fixed adapter@1 & $-0.00434$ & $+0.00387$ & (0.00054, 0.00723) & 0.9892\\
Fixed adapter@6 & $-0.00894$ & $+0.00820$ & (0.00492, 0.01157) & 1.0000\\
Fixed adapter@10 & $-0.01129$ & $+0.01017$ & (0.00703, 0.01343) & 1.0000\\
\bottomrule
\end{tabular}
\begin{tablenotes}[flushleft]
\footnotesize
\item[] Util., mean realized cumulative utility (F1 units, net of costs) for the comparator policy; $\Delta$ mean and 95\% CrI, posterior mean and credible interval of the paired utility difference (HB-PVI minus comparator); $P(\Delta>0)$, posterior probability that HB-PVI's utility was strictly greater. The always-stop comparator is omitted: under the primary setting, HB-PVI purchased zero labels and selected population inference for all 47 participants, so its decisions and realized utility are identical to always-stop by construction (paired difference exactly 0 for every participant; ROPE probability $=1$; see Section~III-D).
\end{tablenotes}
\end{threeparttable}
\end{table}

\begin{figure*}[!t]
\centering
\includegraphics[width=0.92\textwidth]{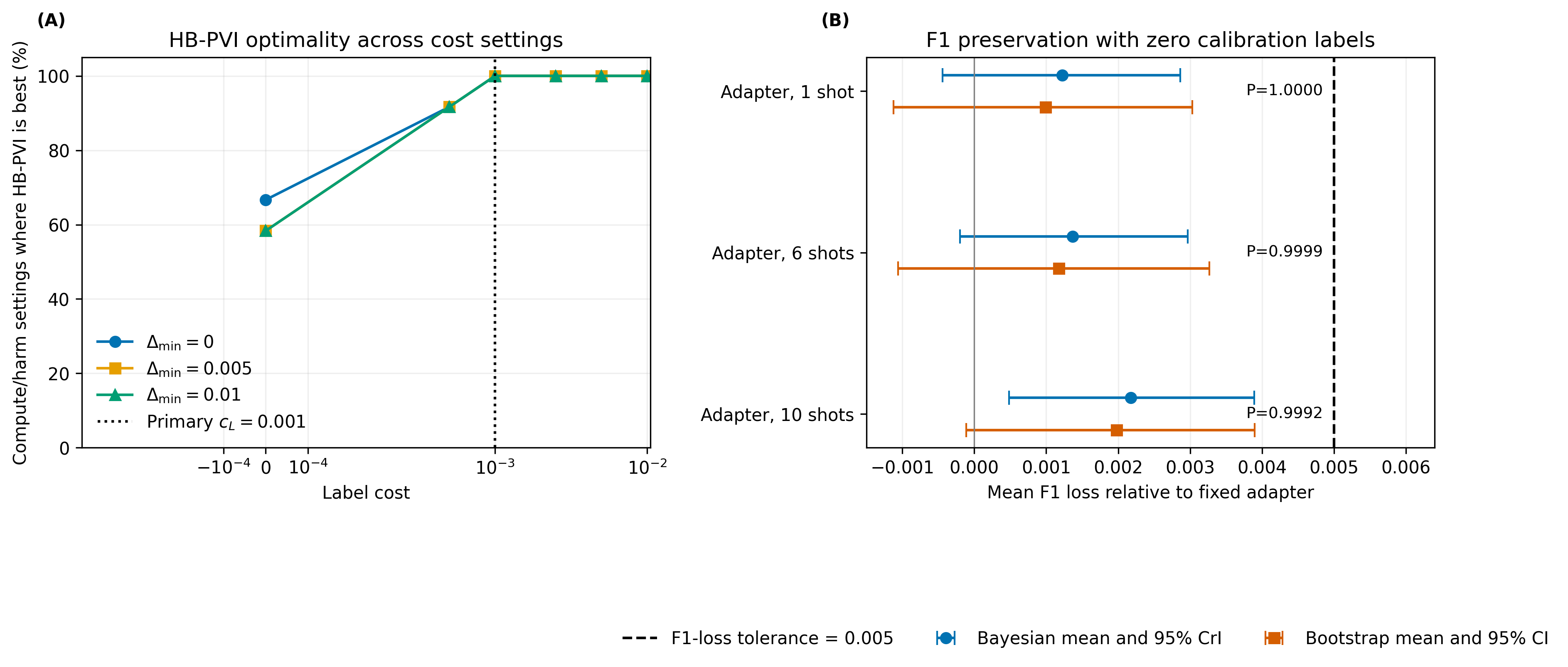}
\caption{Cost sensitivity and F1 preservation with zero calibration labels. (A) Percentage of computation-cost and harm-penalty settings in which HB-PVI was utility-optimal, by label cost and practical-gain threshold; the dotted vertical line marks the primary label cost. (B) Bayesian and bootstrap estimates of mean F1 loss from purchasing zero labels rather than using fixed adapter personalization at 1, 6, or 10 shots. The dashed vertical line marks the F1-loss tolerance of 0.005, and the annotations report the posterior probability that the mean loss remained below this tolerance.}
\label{fig:sensitivity}
\end{figure*}

Across all 216 cost-threshold settings, HB-PVI was utility-optimal in 199 (fixed adapter@10 in 14, fixed adapter@1 in 3). Table~\ref{tab:sensitivity} shows why: at zero label cost, HB-PVI was optimal in only 61.1\% of computation-and-harm-penalty settings, but optimality reached 100\% at and above the primary label cost of $c_L=0.001$, across every value of $\Delta_{\min}$ tested (Fig.~\ref{fig:sensitivity}A).

\begin{table}[!t]
\caption{Percentage of the 36 computation-and-harm-penalty settings, per label cost $c_L$, in which HB-PVI was utility-optimal (across all $\Delta_{\min}$)}
\label{tab:sensitivity}
\centering
\scriptsize
\begin{threeparttable}
\begin{tabular}{lcccccc}
\toprule
$c_L$ & 0 & 0.0005 & 0.001 & 0.0025 & 0.005 & 0.01\\
\midrule
HB-PVI optimal (\%) & 61.1 & 91.7 & 100.0 & 100.0 & 100.0 & 100.0\\
\bottomrule
\end{tabular}
\begin{tablenotes}[flushleft]
\footnotesize
\item[] $c_L$, per-label acquisition cost (F1-equivalent units; Section~II-E). Each column aggregates the 36 combinations of computation cost $c_C$, harm penalty $c_R$, and practical-gain threshold $\Delta_{\min}$ at that label cost.
\end{tablenotes}
\end{threeparttable}
\end{table}

\subsection{Label efficiency and F1 preservation}
%Relative to fixed 10-shot adapter personalization, HB-PVI reduced labeling by 100\% while the Bayesian posterior mean F1 loss was 0.00217 (95\% CrI, 0.00048--0.00389), with $P(\mu_{\mathrm{loss}}<0.005\mid D)=0.9992$; an independent 200{,}000-resample participant bootstrap gave a consistent estimate ($0.00198$; 95\% CI $-0.00011$ to $0.00390$; probability $0.9992$). Both methods therefore agreed that eliminating calibration labeling entirely kept expected F1 loss below a practically negligible 0.005 threshold with high posterior confidence, with probabilities of $1.0000$ and $0.9999$ against the smaller 1- and 6-shot adapter references, respectively (Fig.~\ref{fig:sensitivity}B).

Relative to fixed 10-shot adapter personalization, HB-PVI reduced labeling by 100\% while the Bayesian posterior mean F1 loss was 0.00217 (95\% CrI, 0.00048--0.00389; $P(\mu_{\mathrm{loss}}<0.005\mid D)=0.9992$); a 200{,}000-resample bootstrap gave a similar point estimate (0.00198) but a wider interval that included zero (95\% CI, $-0.00011$ to $0.00390$), reflecting its lack of hierarchical shrinkage across heterogeneous participant-level losses. Both methods agreed the loss stayed below the 0.005 tolerance with probability $\geq0.999$, including $1.0000$ and $0.9999$ against the 1- and 6-shot references (Fig.~\ref{fig:sensitivity}B).

\subsection{Harm avoidance without demonstrated harm identification}
\label{sec:crit3}
At the primary setting, HB-PVI's decision rule selected population inference for all 47 participants, so no participant was exposed to realized personalization harm in this study. This reflects the conservative decision rule under the specified costs, however, rather than a demonstrated ability to identify in advance which participants would be harmed: the fitted gain model's harm-specific Brier score was \emph{worse} than a trivial baseline that always predicts each mechanism's empirical harm rate, for all four personalization mechanisms (Section~\ref{sec:calibration}). Harm was therefore avoided in this study by declining to personalize under the specified costs, not by successfully flagging high-risk participants prospectively -- a distinction we return to in the Discussion.

\subsection{Summary of study objectives}
In summary, HB-PVI eliminated calibration labeling while preserving mean F1 within tolerance, was practically equivalent to (though not demonstrably better than) always-stop, and did not establish prospective discrimination of harmful personalization; participant-level heterogeneity nevertheless changed the deployment recommendation toward a population-first strategy. Full definitions and findings for each study objective are in Supplementary Table~S12.

\section{Discussion}
In a leakage-safe, 47-participant outer evaluation, personalization produced heterogeneous but generally small changes in sample-level F1. Adapter personalization offered the most favorable risk-benefit profile of the four mechanisms tested, with small, monotonically increasing mean gains and comparatively low harm; adapter-plus-head amplified both benefit and harm; and prototype-residual was reliably unfavorable. This matches the intuition that more heavily parameterized mechanisms can both help and hurt a given participant more, reflected directly in the calibration results, where the two more aggressive mechanisms had larger prediction error and below-nominal coverage.

The decision-analytic layer changed the interpretation of these predictive results. Under the primary practical-benefit, safety, and cost assumptions, the expected value of one additional label never exceeded its cost, at any point in the ten-shot sequence, for any participant. The resulting policy purchased zero labels and retained population inference for everyone -- not because personalization was ineffective on average, but because its economic value, once labeling, computation, and harm costs were priced in, was smaller than the price of finding out whether it would help a given participant. This distinction between predictive benefit and decision-relevant value is the central methodological point of HB-PVI: average F1 improvement, the outcome most commonly reported in the HAR personalization literature \cite{phart}, is not by itself sufficient evidence for a deployment decision.

The cost-sensitivity analysis qualifies rather than reverses this conclusion. HB-PVI was utility-optimal in the overwhelming majority of the primary cost-threshold grid and in every setting at or above the primary label cost; fixed personalization was preferable only in a minority of settings with very low label cost and low harm penalty. This indicates the population-first recommendation is a property of the specific cost structure assumed here, not a universal claim that personalization is never worthwhile, and argues for reporting decision surfaces over a cost grid rather than a single point estimate whenever costs are only approximately known.

\subsection{Implications for deployment}
Translated into an operational recommendation, these results argue against building a routine per-user calibration step into a MUSIC-CAR-style deployment under the primary cost assumptions: a new participant can be served directly by the population model, with no adaptation labels collected, while expected F1 loss relative to a hypothetical adapter-personalized deployment stays within tolerance with posterior probability 0.9992. This recommendation is conditional on the representation-learning pipeline, cohort, and cost weights used here, and should be revisited whenever any of three things changes materially. First, if the population backbone is retrained on a smaller or differently sourced cohort and no longer reaches comparably strong F1, the headroom available to personalization enlarges mechanically (Section III-B) and could reverse the decision. Second, if the relative costs of a label, computation, and harm differ from the values used here, Table~\ref{tab:sensitivity} shows the recommendation is not robust at very low label cost. Third, applications placing asymmetric value on avoiding subgroup-specific harm are not directly targeted by the aggregate mean-utility criterion used here and may warrant a subgroup-specific harm constraint instead. None of this requires re-deriving the EVSI framework itself; only the cost inputs to Table~\ref{tab:sensitivity} and the decision-region grid (Supplementary Table~S9) would need recomputing.

\subsection{Limitations}
The cohort contained 47 participants, limiting power for rare subgroups and higher-order interactions. Calibration was weakest for the two more heavily parameterized mechanisms, and harm could not be prospectively identified (Section~\ref{sec:crit3}). The primary EVSI analysis evaluated only the next chronological window, and costs used a common grid rather than mechanism-specific runtime; these weights should be re-elicited for deployment. The label-efficiency result concerns the mean F1 difference, not every individual's loss. Finally, the tested mechanisms were parameter-efficient adaptations of a frozen, population-trained backbone, so larger gains from untested full retraining cannot be ruled out.

\section{Conclusion}
HB-PVI provides a fully Bayesian route from participant-level predictive uncertainty to personalization and label-acquisition decisions for complex activity recognition. Under the primary conditions, no further calibration labels were justified for the MUSIC-CAR cohort, and a population-first policy preserved F1 within tolerance while eliminating calibration labeling entirely. Relative to fixed 10-shot adapter personalization, this policy achieved a 100\% reduction in calibration labels, with posterior probability 0.9992 that the mean F1 loss remained below the 0.005 tolerance. These findings demonstrate that small improvements in average predictive performance do not necessarily provide sufficient decision-relevant value to justify participant-specific adaptation. The framework and its EVSI stopping rule generalize to other wearable-sensing personalization problems in which the benefit of adaptation must be weighed against the cost of acquiring it.

\section*{Data and Code Availability}
The MUSIC-CAR sensor dataset is available from the CARE Lab at Worcester Polytechnic Institute (WPI) at \url{https://carelab.wpi.edu/music_car/}. Data preprocessing and analysis code will be made available on GitHub upon publication.

\section*{Funding}
The author received no financial support for the research.

\section*{Acknowledgment}
The author thanks the MUSIC-CAR data collection team at the CARE Lab at WPI. Results in this paper were obtained using a high-performance computing system acquired through NSF MRI grant DMS-1337943 to WPI.

\bibliographystyle{IEEEtran}
\bibliography{HB-PVI_refs}

\end{document}

\maketitle

This supplement provides tables and methodological detail that support the main manuscript. Section numbers, table numbers, and figure numbers in this document are prefixed with ``S'' and are distinct from those in the main text. Every table below is accompanied by a short statement of what it shows beyond the corresponding main-text result.

\section{Sequential participant-effect updater}
\label{sec:updater}

The main text summarizes the adaptive tempered resample-move sequential Monte Carlo (SMC) algorithm used to update each participant's effect posterior as adaptation labels are purchased. This section gives its full specification, the priors used in Model I and Model II, and the validation gates satisfied on all 47 outer folds.

\subsection{Population posterior used for new-participant updating}
For each outer fold, the Model I hierarchical prediction model was estimated using four NUTS chains, 2,000 warmup iterations per chain, and 4,000 retained draws per chain, yielding 16,000 population-posterior draws containing the activity intercepts, representation coefficients, participant-effect scales, and Cholesky factor of the participant-effect correlation matrix. The outer and validation participants were excluded from population-model fitting.

\subsection{Particle initialization and target distribution}
Let $\bm\Theta^{(s)}$ denote retained population-posterior draw $s=1,\ldots,16{,}000$. For each $\bm\Theta^{(s)}$, four conditionally independent new-participant activity-effect particles are initialized,
\begin{equation}
\mathbf z_{p^\star}^{(s,m)}\sim\mathcal N_9(\mathbf 0,\mathbf I),\qquad
\mathbf b_{p^\star}^{(s,m)}=\operatorname{diag}(\bm\tau_b^{(s)})\mathbf L_{\Omega_b}^{(s)}\mathbf z_{p^\star}^{(s,m)},
\end{equation}
for $m=1,\ldots,4$, so every fold begins with 64,000 joint particles retaining 16,000 distinct population-posterior ancestors. At state $n\in\{0,\ldots,10\}$ the target distribution is proportional to the population posterior, the conditional Gaussian prior for the new-participant effect, and the likelihood of exactly the first $n$ chronological purchased adaptation windows; no future adaptation label is used at state $n$.

\subsection{Adaptive tempered resample-move algorithm}
For each newly purchased window, the algorithm uses adaptive likelihood tempering: if the complete incremental likelihood would reduce particle effective sample size (ESS) below 25\% of 64,000, bisection selects an intermediate likelihood temperature. Systematic resampling is performed only at the 25\% ESS threshold. Following resampling, random-walk Metropolis-Hastings moves rejuvenate the nine-dimensional new-participant activity effect, with the target including the conditional Gaussian prior and the cumulative adaptation likelihood through the current shot; incremental likelihoods are recomputed after every rejuvenation. Population-level posterior parameters are never rejuvenated; consequently every state was required to retain at least 400 distinct population-draw ancestors in addition to a final particle ESS of at least 400. A violation of either gate places the fold on hold and excludes the state from downstream utility modeling. For every state, the algorithm stores posterior means, standard deviations, and 95\% credible intervals for the nine participant effects; posterior predictive activity-probability means and standard deviations; binary predictive entropy; shot-to-shot changes; particle ESS; maximum normalized particle weight; tempering and resampling counts; Metropolis-Hastings acceptance; and the number of distinct population-posterior ancestors.

\subsection{Prior specifications for Model I and Model II}
\label{sec:priors}
Table~\ref{tab:priors} gives the complete prior specification for both models, matching the fitted specification described in the main text.

\begin{table}[h]
\centering
\caption{Prior specifications for Model I and Model II}
\label{tab:priors}
\small
\begin{threeparttable}
\begin{tabular}{p{0.30\linewidth}p{0.64\linewidth}}
\toprule
Parameter & Prior\\
\midrule
\multicolumn{2}{l}{\textit{Model I (hierarchical multi-label prediction)}}\\
$\alpha_k$ (activity intercept) & $\mathcal N(0,2.5^2)$\\
$\beta_{kq}$ (representation coef.) & Regularized horseshoe: $\beta_{kq}=\beta^{\mathrm{raw}}_{kq}\tau_{\beta,k}\tilde\omega_{kq}$, $\tilde\omega_{kq}=c\,\omega_{kq}/\sqrt{c^2+(\tau_{\beta,k}\omega_{kq})^2}$, $c=0.5$\\
$\beta^{\mathrm{raw}}_{kq}$ & $\mathcal N(0,1)$\\
$\omega_{kq}$ (local shrinkage) & $\mathcal N^+(0,1)$\\
$\tau_{\beta,k}$ (global shrinkage) & $\mathcal N^+(0,0.10^2)$\\
$\bm\tau_b$ (participant-effect scale) & $\mathcal N^+(0,0.75^2)$\\
$\bm\Omega_b$ (participant-effect corr.) & $\mathrm{LKJ}(2)$\\
\midrule
\multicolumn{2}{l}{\textit{Model II (hierarchical gain model)}}\\
$\alpha_{a}$ (action intercept) & $\mathcal N(0,1)$\\
$\bm\gamma_{a}$ (predictor coefficients) & $\mathcal N(0,0.35^2)$ (fixed-scale, no shrinkage)\\
$\bm\tau$ (participant-by-action scale) & $\mathcal N^+(0,0.5^2)$\\
$\bm\Omega$ (participant-by-action corr.) & $\mathrm{LKJ}(2)$\\
$\sigma_{a}$ (residual scale) & $\mathcal N^+(0,1)$\\
$\nu$ (Student-$t$ d.f.) & $\mathrm{Gamma}(2,0.1)$, constrained to $[2,50]$\\
\bottomrule
\end{tabular}
\begin{tablenotes}[flushleft]
\footnotesize
\item[] $\mathcal N^+$ denotes a half-normal distribution (support $\geq0$). $k$ indexes activities (Model I) or actions (Model II, denoted $a$); $q$ indexes representation dimensions.
\end{tablenotes}
\end{threeparttable}
\end{table}

Model I's representation coefficients $\bm\beta_k$ use a regularized horseshoe because the $q$-dimensional representation is high-dimensional relative to the per-fold training data, and sparsity-inducing shrinkage helps prevent overfitting the population model to any one representation dimension. Model II's 19-predictor coefficients $\bm\gamma_a$ use a much simpler fixed-scale Gaussian prior with no shrinkage: with 1,880 held-out predictions across 47 folds and only 19 predictors per action, this ratio does not require sparsity-inducing regularization, and a weakly informative fixed-scale prior was judged sufficient.

\subsection{Validation gates}
\label{sec:gates}
This configuration (four participant-effect particles per population draw, 25\% ESS resampling threshold, tempered likelihood incorporation) was applied uniformly to all 47 outer folds. Table~\ref{tab:gates} summarizes the validation gates and the result on the full 47-fold run.

\begin{table}[h]
\centering
\caption{Validation gates for the sequential updater, evaluated on all 47 outer folds}
\label{tab:gates}
\small
\begin{threeparttable}
\begin{tabular}{p{0.56\linewidth}p{0.34\linewidth}}
\toprule
Gate & Result (47/47 folds)\\
\midrule
Minimum distinct population-posterior ancestors per state & $\geq400$: passed\\
Minimum final joint-particle ESS per state & $\geq400$: passed\\
Zero missing direct-state predictor keys & Passed\\
Zero missing participant-effect predictor keys & Passed\\
Enriched-state completeness (517/517 fold-shot states) & Passed\\
\bottomrule
\end{tabular}
\end{threeparttable}
\end{table}

All gates passed on every fold, with no fold requiring exclusion or a fold-specific configuration override. The sequential-updater approximations (bounded likelihood tempering, a fixed 25\% resampling threshold) therefore did not measurably compromise posterior-ancestry diversity anywhere in the cohort, so the Model II gain-model inputs derived from these states (Section~\ref{sec:predictors}) are not confounded by degenerate particle sets for any participant.

\subsection{Enriched sequential-state dataset}
The state representation used by the Model II gain model combines the information available at each shot. The all-fold dataset contains 517 unique states (47 folds $\times$ 11 shots). Representation shift is calculated from the first $n$ outer-adaptation $q$-dimensional principal-component-analysis (PCA) representations relative to the nested-training reference for that fold; at shot 0, representation-shift values are structural zeros with an availability indicator of zero. Calibration labels, outer-test arrays, and utility outcomes are not joined into this state. Adaptation loss is excluded from the primary state because no complete, consistent, leakage-safe action-by-shot definition was identified for all mechanisms and shots.

\section{Sequential-state predictor dictionary}
\label{sec:predictors}
Table~\ref{tab:predictors} lists the 19 predictors used in the Model II hierarchical gain model, together with their information domain.

\begin{table}[h]
\centering
\setlength{\tabcolsep}{16pt}
\caption{Sequential-state predictor dictionary used in Model II}
\label{tab:predictors}
\small
\begin{threeparttable}
\begin{tabular}{ll}
\toprule
Predictor & Domain\\
\midrule
Shot count & Primary predictor\\
Ensemble mutual information (mean) & Ensemble uncertainty\\
Ensemble predictive entropy (mean) & Ensemble uncertainty\\
Ensemble predictive variance (mean) & Ensemble uncertainty\\
Purchased-label cardinality (mean) & Purchased-label summary\\
Purchased-label pattern diversity & Purchased-label summary\\
Purchased-label positive fraction & Purchased-label summary\\
Participant-effect posterior mean (aggregate mean) & Participant-effect summary\\
Participant-effect posterior mean (SD across activities) & Participant-effect summary\\
Participant-effect posterior SD (aggregate mean) & Participant-effect uncertainty\\
Calibration probability mean (aggregate mean) & Calibration-probability summary\\
Calibration probability mean (SD across activities) & Calibration-probability summary\\
Calibration probability SD (aggregate mean) & Calibration-probability uncertainty\\
Calibration predictive entropy (aggregate mean) & Predictive entropy\\
Change in participant-effect mean from previous shot & Change from previous shot\\
Change in calibration probability from previous shot & Change from previous shot\\
Representation shift: mean training Mahalanobis radius & Representation shift\\
Representation shift: centroid $L_2$ distance & Representation shift\\
Representation shift: centroid Mahalanobis distance & Representation shift\\
\bottomrule
\end{tabular}
\begin{tablenotes}[flushleft]
\footnotesize
\item[] All predictors are standardized using training-fold-only moments within each outer split. Three additional columns (population sample F1, action sample F1, gain) are retained as outcomes, not predictors.
\end{tablenotes}
\end{threeparttable}
\end{table}

These 19 predictors span five distinct information domains rather than shot count alone, which is what allows the Model II gain model (main text, Section~II-D) to condition personalization-benefit predictions on each participant's evolving uncertainty, calibration, and representation-shift trajectory. As Table~\ref{tab:priors} shows, Model II's coefficients on these predictors use a fixed-scale, weakly informative Gaussian prior rather than a sparsity-inducing one; with 19 predictors and 1,880 held-out predictions, the predictor-to-observation ratio is modest enough that this was judged sufficient.

\section{EVSI cost specification}
\label{sec:costspec}
Table~\ref{tab:costspec} reproduces the cost specification used throughout the sensitivity analysis, fixed before any EVSI results were inspected.

\begin{table}[h]
\centering
\setlength{\tabcolsep}{16pt}
\caption{EVSI cost specification}
\label{tab:costspec}
\small
\begin{threeparttable}
\begin{tabular}{ll}
\toprule
Quantity & Value(s)\\
\midrule
Practical-gain thresholds $\Delta_{\min}$ & $\{0,\ 0.005,\ 0.01\}$\\
Eligibility constraint & $P(g>\Delta_{\min})\geq 0.90$, $P(g<0)\leq 0.10$\\
Label-cost grid $c_L$ & $\{0,\ 0.0005,\ 0.001,\ 0.0025,\ 0.005,\ 0.01\}$\\
Adapter computation-cost grid $c_C$ & $\{0,\ 0.0005,\ 0.001\}$\\
Population computation cost & $0$ (fixed)\\
Harm-penalty grid $c_R$ & $\{0,\ 0.005,\ 0.01,\ 0.02\}$\\
Sensitivity combinations & $72$ cost combinations $\times$ 3 thresholds $=216$\\
Primary setting & $c_L=0.001$, $c_C=0.0005$, $c_R=0.01$\\
EVSI cost-free utility & $U'(a,g)=g-c_C(a)-c_R\,\mathbb I(g<0)$\\
Continuation rule & Continue iff $\mathrm{EVSI}_{n+1}>c_L$\\
Label cost charged & Exactly once per purchased label\\
Window order & Fixed chronological (active selection deferred)\\
\bottomrule
\end{tabular}
\end{threeparttable}
\end{table}

This grid is deliberately wide relative to the primary setting so that Table~V and Fig.~4 of the main text can show \emph{how far} the population-first recommendation extends before it is reversed, rather than reporting robustness at a single arbitrary cost. The primary values were fixed before any EVSI result was computed, so the primary-setting conclusion in the main text is not the result of selecting a favorable cost combination after the fact.

\section{Full ten-shot action-specific outcomes, including mean F1}
\label{sec:actionsfull}
Table~\ref{tab:actionsfull} extends Table~I of the main text to all ten purchased-label counts for all four personalization mechanisms, reporting the realized mean sample-level F1 (with its 95\% interval) alongside gain relative to the population baseline (F1 $=0.9042$).

\begin{table}[h]
\centering
\caption{Action-specific mean F1 and gain by shot (all 10 shots; percentages of 47 participants)}
\label{tab:actionsfull}
\small
\begin{threeparttable}
\begin{tabular}{llcccc}
\toprule
Action & Shot & Mean F1 (95\% CI) & Gain & Ben.$>$.01 (\%) & Harm$<-$.01 (\%)\\
\midrule
\multirow{10}{*}{Adapter}
 & 1 & 0.9052 (0.8810, 0.9272) & $+0.00099$ & 8.5 & 4.3\\
 & 2 & 0.9053 (0.8792, 0.9271) & $+0.00109$ & 8.5 & 4.3\\
 & 3 & 0.9053 (0.8797, 0.9270) & $+0.00105$ & 10.6 & 6.4\\
 & 4 & 0.9054 (0.8810, 0.9274) & $+0.00121$ & 8.5 & 4.3\\
 & 5 & 0.9055 (0.8807, 0.9270) & $+0.00124$ & 8.5 & 4.3\\
 & 6 & 0.9054 (0.8808, 0.9262) & $+0.00118$ & 8.5 & 4.3\\
 & 7 & 0.9054 (0.8793, 0.9271) & $+0.00118$ & 10.6 & 4.3\\
 & 8 & 0.9055 (0.8797, 0.9269) & $+0.00133$ & 10.6 & 4.3\\
 & 9 & 0.9061 (0.8810, 0.9280) & $+0.00189$ & 10.6 & 2.1\\
 & 10 & 0.9062 (0.8807, 0.9275) & $+0.00198$ & 10.6 & 4.3\\
\midrule
\multirow{10}{*}{Head}
 & 1 & 0.9053 (0.8785, 0.9278) & $+0.00109$ & 14.9 & 8.5\\
 & 2 & 0.9050 (0.8778, 0.9268) & $+0.00075$ & 14.9 & 10.6\\
 & 3 & 0.9046 (0.8781, 0.9264) & $+0.00038$ & 14.9 & 10.6\\
 & 4 & 0.9044 (0.8779, 0.9270) & $+0.00020$ & 14.9 & 10.6\\
 & 5 & 0.9045 (0.8779, 0.9266) & $+0.00027$ & 14.9 & 10.6\\
 & 6 & 0.9040 (0.8775, 0.9261) & $-0.00019$ & 14.9 & 12.8\\
 & 7 & 0.9041 (0.8778, 0.9262) & $-0.00009$ & 14.9 & 10.6\\
 & 8 & 0.9042 (0.8781, 0.9269) & $-0.00003$ & 17.0 & 12.8\\
 & 9 & 0.9044 (0.8783, 0.9257) & $+0.00021$ & 17.0 & 10.6\\
 & 10 & 0.9046 (0.8782, 0.9263) & $+0.00037$ & 19.1 & 10.6\\
\midrule
\multirow{10}{*}{Adapter+head}
 & 1 & 0.9019 (0.8759, 0.9251) & $-0.00236$ & 17.0 & 21.3\\
 & 2 & 0.9008 (0.8730, 0.9245) & $-0.00341$ & 14.9 & 27.7\\
 & 3 & 0.9000 (0.8719, 0.9232) & $-0.00421$ & 12.8 & 31.9\\
 & 4 & 0.8999 (0.8721, 0.9234) & $-0.00432$ & 12.8 & 29.8\\
 & 5 & 0.8999 (0.8723, 0.9233) & $-0.00433$ & 14.9 & 27.7\\
 & 6 & 0.8995 (0.8728, 0.9224) & $-0.00469$ & 19.1 & 29.8\\
 & 7 & 0.8995 (0.8724, 0.9222) & $-0.00475$ & 19.1 & 29.8\\
 & 8 & 0.8993 (0.8721, 0.9220) & $-0.00492$ & 17.0 & 29.8\\
 & 9 & 0.8997 (0.8735, 0.9230) & $-0.00455$ & 17.0 & 25.5\\
 & 10 & 0.8997 (0.8724, 0.9227) & $-0.00449$ & 21.3 & 25.5\\
\midrule
\multirow{10}{*}{Prototype-residual}
 & 1 & 0.9019 (0.8775, 0.9236) & $-0.00235$ & 0.0 & 8.5\\
 & 2 & 0.9018 (0.8777, 0.9235) & $-0.00246$ & 0.0 & 8.5\\
 & 3 & 0.9017 (0.8775, 0.9229) & $-0.00253$ & 0.0 & 6.4\\
 & 4 & 0.9016 (0.8768, 0.9232) & $-0.00259$ & 0.0 & 6.4\\
 & 5 & 0.9017 (0.8767, 0.9235) & $-0.00247$ & 0.0 & 6.4\\
 & 6 & 0.9015 (0.8758, 0.9228) & $-0.00270$ & 0.0 & 4.3\\
 & 7 & 0.9017 (0.8766, 0.9229) & $-0.00249$ & 0.0 & 6.4\\
 & 8 & 0.9017 (0.8771, 0.9232) & $-0.00255$ & 0.0 & 6.4\\
 & 9 & 0.9015 (0.8759, 0.9224) & $-0.00269$ & 0.0 & 6.4\\
 & 10 & 0.9015 (0.8764, 0.9237) & $-0.00272$ & 0.0 & 6.4\\
\bottomrule
\end{tabular}
\begin{tablenotes}[flushleft]
\footnotesize
\item[] Population-only reference: mean F1 $=0.9042$ (constant across shots and mechanisms by construction, since gain is defined relative to this fixed baseline).
\end{tablenotes}
\end{threeparttable}
\end{table}

Read alongside Table~I of the main text, this table shows that adapter's advantage over the other three mechanisms is present, if small, at every one of the ten shots, while adapter-plus-head's disadvantage widens steadily from shot 1 through roughly shot 8 before leveling off. None of the four mechanisms' mean F1 ever separates from the 0.9042 population baseline by more than about 1.5 percentage points in either direction at the group level -- the large individual-level swings reported in Section~\ref{sec:heterogeneity} occur despite, not because of, the group-level trajectories shown here.

\section{Participant-level heterogeneity in realized adapter gain and exact F1}
\label{sec:heterogeneity}
Table~\ref{tab:heterogeneity} reports the full distribution of realized adapter gain across the 47 participants at 1, 6, and 10 purchased labels.

\begin{table}[h]
\centering
\caption{Distribution of realized adapter gain across 47 participants, by shot count}
\label{tab:heterogeneity}
\small
\begin{threeparttable}
\begin{tabular}{lccccc}
\toprule
Shot & Min & Q1 & Median & Q3 & Max\\
\midrule
1 & $-0.02188$ & $-0.00168$ & $+0.00125$ & $+0.00369$ & $+0.02083$\\
6 & $-0.02486$ & $-0.00214$ & $+0.00164$ & $+0.00469$ & $+0.01932$\\
10 & $-0.02650$ & $-0.00058$ & $+0.00203$ & $+0.00533$ & $+0.01678$\\
\bottomrule
\end{tabular}
\begin{tablenotes}[flushleft]
\footnotesize
\item[] Values are realized gain (action F1 minus population F1) for the fixed-adapter comparator at each shot count, at the primary cost and threshold setting ($\Delta_{\min}=0.01$, $c_L=0.001$, $c_C=0.0005$, $c_R=0.01$); $n=47$ at each shot count.
\end{tablenotes}
\end{threeparttable}
\end{table}

Table~\ref{tab:perparticipant} reports exact participant-specific scores taken directly from the source population and action F1 columns; no participant score is reconstructed from the cohort mean. The population baseline is participant-specific, and the identity $F1_{p,a,n}-F1_{p,\mathrm{population}}=g_{p,a,n}$ was verified for every action-shot row.

\begin{ThreePartTable}
\begin{TableNotes}[flushleft]
\footnotesize
\item[] Fold, anonymized participant identifier (outer-fold index); Population F1, sample-level F1 with no personalization; Adapter F1@1/@6/@10, sample-level F1 after adapter personalization at 1, 6, and 10 purchased labels; Gain@10, Adapter F1@10 minus Population F1.
\end{TableNotes}
\begin{longtable}{rrrrrr}
\caption{Exact participant-specific population and adapter F1 scores, all 47 participants}\label{tab:perparticipant}\\
\toprule
Fold & Population F1 & Adapter F1@1 & Adapter F1@6 & Adapter F1@10 & Gain@10\\
\midrule
\endfirsthead
\multicolumn{6}{c}{Table~\ref{tab:perparticipant} continued}\\
\toprule
Fold & Population F1 & Adapter F1@1 & Adapter F1@6 & Adapter F1@10 & Gain@10\\
\midrule
\endhead
\bottomrule
\insertTableNotes
\endlastfoot
000 & 0.5664 & 0.5606 & 0.5621 & 0.5777 & +0.0113 \\
001 & 0.7462 & 0.7505 & 0.7471 & 0.7482 & +0.0020 \\
002 & 0.6804 & 0.6811 & 0.6772 & 0.6699 & -0.0105 \\
003 & 0.7883 & 0.7887 & 0.7902 & 0.7885 & +0.0002 \\
004 & 0.7906 & 0.7687 & 0.7669 & 0.7641 & -0.0265 \\
005 & 0.9203 & 0.9158 & 0.9183 & 0.9203 & +0.0000 \\
006 & 0.9307 & 0.9239 & 0.9283 & 0.9288 & -0.0018 \\
007 & 0.9746 & 0.9687 & 0.9679 & 0.9716 & -0.0030 \\
008 & 0.9079 & 0.9102 & 0.9043 & 0.9013 & -0.0066 \\
009 & 0.9496 & 0.9465 & 0.9450 & 0.9468 & -0.0029 \\
010 & 0.9568 & 0.9599 & 0.9590 & 0.9594 & +0.0026 \\
011 & 0.9454 & 0.9449 & 0.9478 & 0.9480 & +0.0026 \\
012 & 0.9207 & 0.9193 & 0.9209 & 0.9209 & +0.0001 \\
013 & 0.9163 & 0.9195 & 0.9211 & 0.9211 & +0.0048 \\
014 & 0.8848 & 0.9056 & 0.9041 & 0.9011 & +0.0163 \\
015 & 0.8384 & 0.8468 & 0.8457 & 0.8486 & +0.0103 \\
016 & 0.8849 & 0.8830 & 0.8859 & 0.8833 & -0.0016 \\
017 & 0.9536 & 0.9326 & 0.9287 & 0.9476 & -0.0060 \\
018 & 0.9776 & 0.9788 & 0.9776 & 0.9779 & +0.0003 \\
019 & 0.9617 & 0.9574 & 0.9590 & 0.9620 & +0.0003 \\
020 & 0.8749 & 0.8880 & 0.8847 & 0.8849 & +0.0100 \\
021 & 0.9551 & 0.9566 & 0.9584 & 0.9584 & +0.0033 \\
022 & 0.9386 & 0.9403 & 0.9407 & 0.9421 & +0.0035 \\
023 & 0.9192 & 0.9267 & 0.9245 & 0.9264 & +0.0072 \\
024 & 0.9475 & 0.9507 & 0.9527 & 0.9521 & +0.0046 \\
025 & 0.9561 & 0.9591 & 0.9593 & 0.9582 & +0.0022 \\
026 & 0.9479 & 0.9475 & 0.9488 & 0.9495 & +0.0015 \\
027 & 0.8535 & 0.8652 & 0.8681 & 0.8703 & +0.0168 \\
028 & 0.9634 & 0.9655 & 0.9605 & 0.9637 & +0.0003 \\
029 & 0.9325 & 0.9399 & 0.9372 & 0.9359 & +0.0033 \\
030 & 0.8980 & 0.9024 & 0.9035 & 0.9035 & +0.0054 \\
031 & 0.8552 & 0.8560 & 0.8539 & 0.8538 & -0.0014 \\
032 & 0.9604 & 0.9602 & 0.9599 & 0.9602 & -0.0002 \\
033 & 0.9538 & 0.9572 & 0.9564 & 0.9547 & +0.0010 \\
034 & 0.9002 & 0.9026 & 0.8986 & 0.9054 & +0.0052 \\
035 & 0.9188 & 0.9122 & 0.9177 & 0.9179 & -0.0009 \\
036 & 0.9861 & 0.9858 & 0.9831 & 0.9831 & -0.0030 \\
037 & 0.9582 & 0.9552 & 0.9631 & 0.9631 & +0.0050 \\
038 & 0.9556 & 0.9491 & 0.9508 & 0.9479 & -0.0077 \\
039 & 0.9638 & 0.9665 & 0.9655 & 0.9642 & +0.0004 \\
040 & 0.9569 & 0.9574 & 0.9607 & 0.9626 & +0.0057 \\
041 & 0.8951 & 0.8948 & 0.8979 & 0.9009 & +0.0058 \\
042 & 0.9642 & 0.9646 & 0.9646 & 0.9646 & +0.0004 \\
043 & 0.9611 & 0.9669 & 0.9634 & 0.9642 & +0.0031 \\
044 & 0.9513 & 0.9574 & 0.9585 & 0.9599 & +0.0086 \\
045 & 0.8304 & 0.8451 & 0.8433 & 0.8394 & +0.0090 \\
046 & 0.8052 & 0.8092 & 0.8205 & 0.8170 & +0.0118 \\
\end{longtable}
\end{ThreePartTable}

The exact participant-level results show that the small mean adapter gain at 10 shots ($+0.00198$) masks substantial heterogeneity. Five participants had gains above $+0.01$, two had gains below $-0.01$, and the remaining participants were distributed around smaller positive and negative changes. Because baseline F1 varied substantially across participants (0.566 to 0.986), exact participant-specific scores are more informative than adding gains to the cohort mean, and are the source of the participant-level statements and the population-vs-personalized F1 figure in the main text.

\section{Held-out gain-model calibration diagnostics}
\label{sec:calibsupp}
\begin{figure}[h]
\centering
\includegraphics[width=\linewidth]{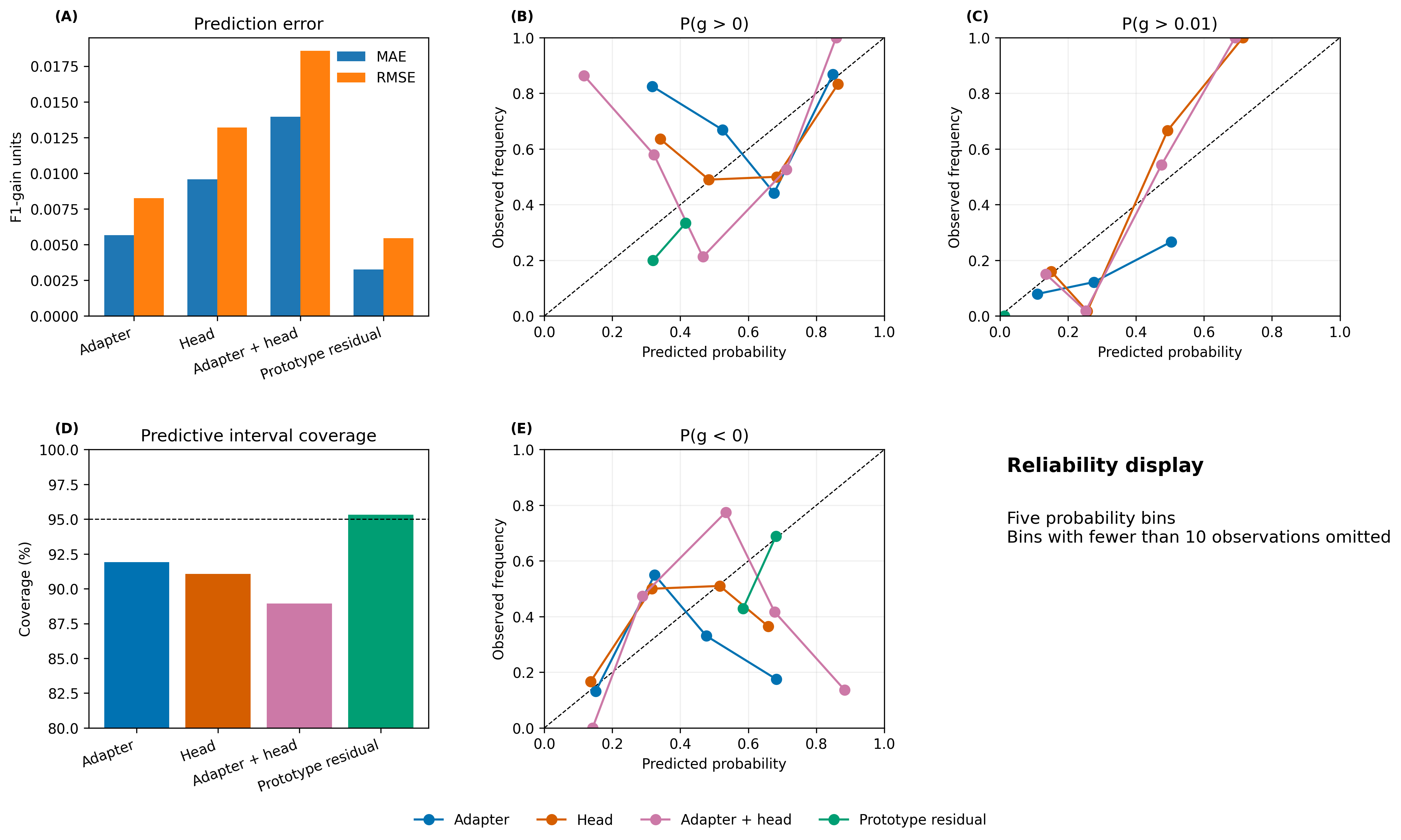}
\caption{Held-out gain-model prediction and calibration. (A) Prediction error (MAE, RMSE) by action. (B–C, E) Reliability curves comparing predicted and observed frequencies for positive gain, gain $>0.01$, and negative gain (five probability bins per action-event combination; bins with $<10$ held-out observations omitted). (D) Empirical coverage of the nominal 95\% predictive interval by action, with the 95\% reference line.}
\label{fig:s1}
\end{figure}

Table~III of the main text summarizes held-out prediction error and calibration in numeric form. Fig.~\ref{fig:s1} shows the underlying reliability curves and per-action breakdown, which reveal \emph{where} each mechanism's miscalibration concentrates rather than only its overall magnitude.

Two patterns stand out beyond the aggregate numbers in Table~III. First, in panel (C), adapter's meaningful-benefit reliability curve stays well below the diagonal across almost its entire range, meaning the model consistently assigns a higher probability of meaningful benefit than is actually observed for this mechanism, even though adapter has the lowest overall prediction error of the four actions. Second, and more consequential, panel (E) shows that adapter-plus-head's harm-probability curve is not merely miscalibrated but inverts at the extreme: the bin with the \emph{highest} predicted harm probability ($\approx0.88$) has the \emph{lowest} observed harm frequency ($\approx0.13$) of any bin for that action -- the opposite of what a well-calibrated model would show. This is the clearest single illustration of the harm-identification result in the main text (Section~III-H): even the most confident harm predictions for the most heavily parameterized mechanism cannot be trusted directionally, not only imprecisely.

\section{Full paired-comparison posterior diagnostics}
\label{sec:paired}
Table~\ref{tab:paireddiag} extends Table~IV of the main text with the underlying MCMC diagnostics for the participant-level Student-$t$ paired-utility-difference models (HB-PVI minus each comparator, primary setting).

\begin{table}[h]
\centering
\caption{Paired-comparison posterior diagnostics (HB-PVI minus comparator)}
\label{tab:paireddiag}
\small
\begin{threeparttable}
\begin{tabular}{lccccc}
\toprule
Comparator & Div. & Max treedepth & Max $\widehat R$ & Min bulk ESS & Min tail ESS\\
\midrule
Always-stop & --- & --- & --- & --- & ---\tnote{a}\\
Fixed adapter@1 & 0 & 0 & 1.0012 & 5871 & 5802\\
Fixed adapter@6 & 0 & 0 & 1.0007 & 5224 & 5909\\
Fixed adapter@10 & 0 & 0 & 1.0008 & 4099 & 3495\\
\bottomrule
\end{tabular}
\begin{tablenotes}[flushleft]
\footnotesize
\item[a] All 47 paired differences are exactly zero; MCMC not applicable (handled analytically).
\item[] Div., number of divergent transitions; Max treedepth, number of iterations hitting the maximum treedepth; $\widehat R$, rank-normalized potential scale reduction factor; ESS, effective sample size. All non-degenerate comparisons pass the convergence gates ($\widehat R<1.01$, zero divergences, zero max-treedepth hits, bulk/tail ESS$>400$).
\end{tablenotes}
\end{threeparttable}
\end{table}

All three non-degenerate models converged cleanly with no divergences and $\widehat R\leq1.0012$, so the posterior paired-utility-difference estimates in Table~IV of the main text (and the corresponding $P(\mu_d>0)$ values) are not confounded by sampler pathologies; the 10-shot comparison had the smallest ESS among the three, but its minimum ESS remained above 3,400, with no divergences and maximum $\widehat R<1.01$.

\section{Decision-region breakdown by threshold and label cost}
\label{sec:decisionregion}
Table~\ref{tab:decisionregion} gives the full 18-row breakdown (three practical-gain thresholds $\times$ six label costs, each aggregating 12 computation-cost/harm-penalty combinations) underlying Fig.~4A and Table~V of the main text.

\begin{table}[h]
\centering
\setlength{\tabcolsep}{18pt}
\small
\begin{threeparttable}
\caption{HB-PVI utility-optimality by practical-gain threshold and label cost}
\label{tab:decisionregion}
\begin{tabular}{cccc}
\toprule
$\Delta_{\min}$ & $c_L$ & HB-PVI best (\%) & Runner-up policy (count)\\
\midrule
0.000 & 0.0000 & 66.7 & fixed\_adapter\_10 (4)\\
0.000 & 0.0005 & 91.7 & fixed\_adapter\_1 (1)\\
0.000 & 0.0010 & 100.0 & ---\\
0.000 & 0.0025 & 100.0 & ---\\
0.000 & 0.0050 & 100.0 & ---\\
0.000 & 0.0100 & 100.0 & ---\\
0.005 & 0.0000 & 58.3 & fixed\_adapter\_10 (5)\\
0.005 & 0.0005 & 91.7 & fixed\_adapter\_1 (1)\\
0.005 & 0.0010 & 100.0 & ---\\
0.005 & 0.0025 & 100.0 & ---\\
0.005 & 0.0050 & 100.0 & ---\\
0.005 & 0.0100 & 100.0 & ---\\
0.010 & 0.0000 & 58.3 & fixed\_adapter\_10 (5)\\
0.010 & 0.0005 & 91.7 & fixed\_adapter\_1 (1)\\
0.010 & 0.0010 & 100.0 & ---\\
0.010 & 0.0025 & 100.0 & ---\\
0.010 & 0.0050 & 100.0 & ---\\
0.010 & 0.0100 & 100.0 & ---\\
\bottomrule
\end{tabular}
\begin{tablenotes}[flushleft]
\footnotesize
\item Each row aggregates 12 computation-cost/harm-penalty settings.
\end{tablenotes}
\end{threeparttable}
\end{table}

\begin{figure}[h]
\centering
\includegraphics[width=\linewidth]{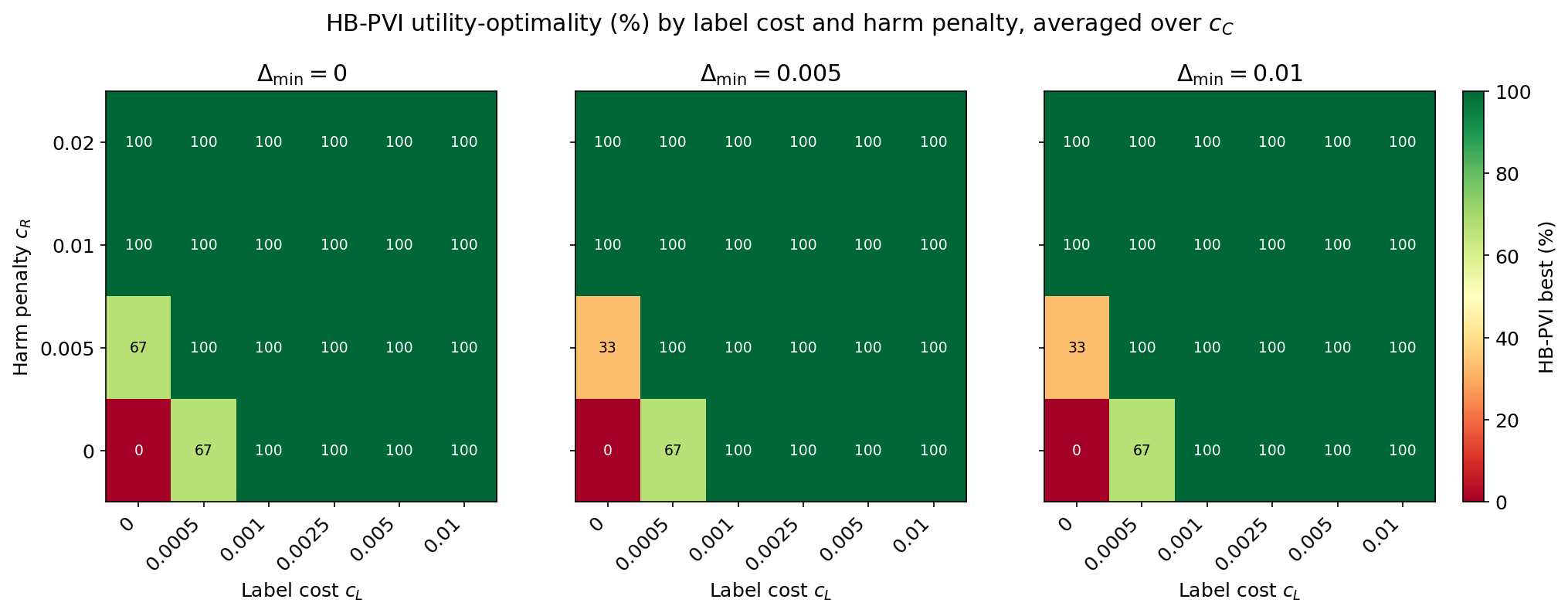}
\caption{HB-PVI utility-optimality (\%) across the full label-cost ($c_L$) and harm-penalty ($c_R$) grid, averaged over the three computation-cost ($c_C$) settings, for each practical-gain threshold. HB-PVI is never preferred only in the single cell where both costs are zero; it is preferred in a minority of settings only when exactly one of the two costs is zero and the other is at its lowest tested nonzero level; everywhere else in the grid, HB-PVI is optimal in 100\% of cases.}
\label{fig:s2}
\end{figure}

The 100.0\% rows are the state underlying the primary-setting result in the main text: once label cost reaches 0.001, HB-PVI wins in every one of the 12 computation-cost/harm-penalty combinations at every practical-gain threshold tested, not just on average across them. The only settings in which a fixed-shot comparator ever wins are at label cost 0 or 0.0005, an order of magnitude below the primary value. This is consistent with Section~\ref{sec:stopfreq}'s finding that the largest one-step EVSI observed anywhere in the study (0.01012) narrowly exceeds only the two smallest label costs in the grid, and that positive EVSI was confined throughout to $\Delta_{\min}=0$ and a single participant. Fig.~\ref{fig:s2} resolves this label-cost margin into its two underlying cost dimensions by also varying the harm penalty $c_R$.

Fig.~\ref{fig:s2} shows that the reversal region identified in Table~\ref{tab:decisionregion} is smaller than the label-cost margin alone suggests: HB-PVI loses to a fixed-shot comparator in only three of the 24 cells in each panel (0\%, 66.7\%, and either 66.7\% or 33.3\% depending on $\Delta_{\min}$, all confined to $c_L\leq0.0005$ combined with $c_R\leq0.005$), and is optimal in 100\% of settings as soon as either cost moves to its next tested level. Harm penalty and label cost therefore behave as substitutes in this decision: a moderate increase in either one is enough to restore the population-first recommendation even while the other remains at its lowest tested value.

\section{Label-efficiency Bayesian and bootstrap diagnostics}
\label{sec:crit2}
Table~\ref{tab:crit2diag} reports the posterior degrees-of-freedom and scale parameters and convergence diagnostics for the Bayesian label-efficiency model underlying Section~III-G of the main text.

\begin{table}[h]
\centering
\caption{Label-efficiency posterior diagnostics (Bayesian Student-$t$ model)}
\label{tab:crit2diag}
\small
\begin{threeparttable}
\begin{tabular}{lccccc}
\toprule
Reference & Post.\ $\tau$ & Post.\ $\nu$ & Max $\widehat R$ & Min bulk ESS & Min tail ESS\\
\midrule
Fixed adapter@1 & 0.00490 & 4.52 & 1.0011 & 3001 & 3300\\
Fixed adapter@6 & 0.00468 & 3.70 & 1.0009 & 4014 & 4088\\
Fixed adapter@10 & 0.00510 & 5.75 & 1.0007 & 4159 & 4349\\
\bottomrule
\end{tabular}
\begin{tablenotes}[flushleft]
\footnotesize
\item[] Post.\ $\tau$ and post.\ $\nu$ are the posterior-mean scale and degrees-of-freedom of the participant-level Student-$t$ F1-loss model; other columns as in Table~\ref{tab:paireddiag}.
\end{tablenotes}
\end{threeparttable}
\end{table}

Convergence was clean for all three reference comparisons (zero divergences, $\widehat R\leq1.0011$), so the posterior probabilities reported in the main text (e.g., $P(\mu_{\mathrm{loss}}<0.005\mid D)=0.9992$ against the 10-shot reference) rest on well-mixed chains rather than an under-explored posterior. The low degrees-of-freedom estimates ($\nu\approx3.7$--$5.8$) indicate a genuinely heavy-tailed distribution of participant-level F1 losses, consistent with the extreme individual cases identified in Table~\ref{tab:perparticipant}, and justify the use of a Student-$t$ rather than Gaussian likelihood for this comparison.

\section{Stopping-decision and action-selection frequencies}
\label{sec:stopfreq}
Across the complete 47-fold $\times$ 216-setting stopping-decision dataset (10,152 fold-setting decisions), 10,116 (99.6\%) stopped because one-step EVSI did not exceed the label cost, and the remaining 36 (0.4\%) stopped only because the ten-label horizon was reached (all occurring at $\Delta_{\min}=0$ with $c_L=0$). Population inference was selected in 10,080 of 10,152 decisions (99.3\%); head personalization was selected in the remaining 72 decisions (0.7\%), all at $\Delta_{\min}=0$. Adapter, adapter-plus-head, and prototype-residual personalization were never selected as the terminal action under any cost-threshold setting, consistent with the primary-setting result reported in the main text.

Table~\ref{tab:evsiceiling} details the 72 decisions with strictly positive one-step EVSI referenced in Section~III-E of the main text, grouped by label cost (each group aggregates 12 computation-cost/harm-penalty combinations, all at $\Delta_{\min}=0$ and all attributable to the same single participant).

\begin{table}[h]
\centering
\setlength{\tabcolsep}{16pt}
\small
\begin{threeparttable}
\caption{Positive one-step EVSI decisions, grouped by label cost (all at $\Delta_{\min}=0$; $n=12$ decisions per row)}
\label{tab:evsiceiling}
\begin{tabular}{cccc}
\toprule
$c_L$ & Mean EVSI at stop & Max EVSI observed & Mean labels purchased\\
\midrule
0.0000 & 0.00405 & 0.00411 & 10.0\\
0.0005 & 0.00405 & 0.00411 & 10.0\\
0.0010 & 0.00405 & 0.00411 & 10.0\\
0.0025 & 0.00956 & 0.01012 & 1.0\\
0.0050 & 0.00956 & 0.01012 & 1.0\\
0.0100 & 0.00956 & 0.01012 & 1.0\\
\bottomrule
\end{tabular}
\end{threeparttable}
\end{table}

The same one participant accounts for every positive-EVSI decision in the entire 10,152-decision grid. Their behavior under cost changes is itself informative: at low label cost the model judges it worthwhile to keep purchasing all the way to the ten-label horizon (mean EVSI per continued step, 0.0041), but once label cost exceeds about 0.002 the model buys exactly one label and stops, because EVSI drops sharply after the first purchase. This is the clearest illustration in the study of EVSI behaving as intended -- responding to both the participant's own trajectory and the price of information -- even though, in aggregate, this single case is not enough to change the population-first recommendation for the cohort as a whole.

\section{Study objectives and outcomes}
\label{sec:objectives}
The study's five objectives were specified before any Model~I/II fitting or EVSI computation. Table~\ref{tab:criteria} summarizes each objective and its outcome, drawing on results reported throughout the main text and this supplement.

\begin{table}[h]
\centering
\caption{Study objectives and outcomes}
\label{tab:criteria}
\small
\begin{threeparttable}
\begin{tabular}{p{0.05\linewidth}p{0.60\linewidth}p{0.23\linewidth}}
\toprule
\# & Objective & Status\\
\midrule
1 & $P$(mean utility $>$ best existing policy)$>0.95$ without worse calibration & Not met\\
2 & $\geq$25\% fewer labels with $P$(F1 loss$<$0.005)$>0.90$ & Met\\
3 & Materially improved prospective harmful-personalization identification & Not established\\
4 & Practical equivalence supporting simpler policy, ROPE $P>0.90$ & Met (ROPE $P=1$)\\
5 & Heterogeneity estimate changes deployment recommendation & Supported\\
\bottomrule
\end{tabular}
\begin{tablenotes}[flushleft]
\footnotesize
\item[] ROPE, region of practical equivalence (main text, Section~II-F).
\end{tablenotes}
\end{threeparttable}
\end{table}

Objective 1 was not met because HB-PVI was exactly equivalent to, rather than better than, always-stop under the primary cost setting (main text, Section~III-F) -- a finding that itself supports objective 4 (practical equivalence, met exactly with ROPE probability 1). Objective 2 was met with high posterior confidence by both a Bayesian and an independent bootstrap analysis (Section~\ref{sec:crit2} above; main text, Section~III-G). Objective 3 was not established: although the primary decision rule avoided realized harm by selecting population inference for everyone, the underlying harm-probability predictions did not discriminate risk better than a trivial baseline (main text, Sections~III-C and III-H). Objective 5 was supported by the substantial participant-level heterogeneity documented in Tables~\ref{tab:heterogeneity} and~\ref{tab:perparticipant} above, which shows that a small subset of participants would benefit or be harmed considerably more than the population-averaged effect suggests, even though this heterogeneity could not be predicted early enough to justify acting on it under the specified costs.